\documentclass[10pt,twocolumn,letterpaper]{article}

\usepackage{iccv}

\usepackage[utf8]{inputenc} 
\usepackage[T1]{fontenc}    
\usepackage{url}            
\usepackage{booktabs}       
\usepackage{amsfonts}       
\usepackage{nicefrac}       
\usepackage{microtype}      
\usepackage{xcolor}         
\usepackage{subcaption}

\usepackage{makecell}
\usepackage{multirow}
\usepackage{color}
\usepackage{graphicx}
\definecolor{DeepPink}{RGB}{255,20,147}
\usepackage{booktabs, multirow}
\usepackage{xcolor,amsmath}
\definecolor{darkblue}{rgb}{0,0.08,0.45}
\definecolor{cvprblue}{rgb}{0.21,0.49,0.74}
\definecolor{mygreen2}{RGB}{0 205 0}
\usepackage{colortbl}
\usepackage{amsfonts,bm,pifont}
\usepackage{wrapfig}
\usepackage{marvosym}
\usepackage{placeins}
\usepackage[pagebackref,breaklinks,colorlinks]{hyperref}

\title{VLA-Thinker: Boosting Vision-Language-Action Models through Thinking-with-Image Reasoning}

\author{%
  Chaoyang Wang$^{1}$\quad 
  Wenrui Bao$^{1}$\footnotemark[\value{footnote}]\quad 
  Sicheng Gao$^{2}$\quad Bingxin Xu$^{3}$  \quad Yu Tian$^{1}$\\
 Yogesh S Rawat$^{1}$\quad Yunhao Ge$^{4}$\quad Yuzhang Shang$^{1}$\thanks{Corresponding Author.}\\ [0.8ex]
  $^{1}$University of Central Florida \quad $^{2}$University of Würzburg \quad $^{3}$University of Southern California \\
  $^{4}$NVIDIA Research\\
    \tt\href{https://cywang735.github.io/VLA-Thinker/}{Project\&Codes:VLA-Thinker}
    \vspace{-0.2in}
}

\begin{document}

\maketitle

\vspace{-0.2in}
\begin{abstract}
Enabling Vision–Language–Action (VLA) models to ``think before acting'' via Chain-of-Thought (CoT) reasoning has emerged as a promising direction for improving data efficiency and decision robustness in embodied intelligence. However, existing CoT-enhanced VLA approaches remain constrained by a text-based paradigm: visual observations are encoded once as static context, while reasoning unfolds primarily in the language space. Such a design limits cross-modal interaction and prevents the model from actively revisiting the environment to resolve ambiguities or recover from intermediate errors, particularly in long-horizon manipulation tasks.
To address these challenges, we propose VLA-Thinker, a thinking-with-image reasoning framework for embodied intelligence that aims to break away from text-based chain-of-thought reasoning by treating visual perception as an explicit component of the reasoning process. 
Unlike traditional VLA approaches that regard visual input as a one-shot observation, VLA-Thinker actively acquires task-relevant visual information through tool invocation during reasoning, thereby enabling an interleaved and cooperative perception–reasoning–action process.
Training such a system, however, presents unique challenges: the model must learn not only what to reason, but when and how to query visual information, and how to align complete reasoning–action trajectories with task success. To this end, we introduce a two-stage pipeline:
(1) a SFT cold start phase using carefully curated visual CoT data to distill foundational reasoning capabilities and operation formats; and (2) the application of Group Relative Policy Optimization (GRPO) to causally align the complete reasoning–action trajectories with desired task outcomes.
Experimental results demonstrate that VLA-Thinker achieves  significant performance improvements on both the LIBERO (97.5\%) and the RoboTwin 2.0 (62.3\%, 70.7\%, and 64.6\%).
\end{abstract}

\vspace{-0.3in}
\section{Introduction}
\label{sec:intro}

Vision-Language-Action (VLA) models have emerged as a promising paradigm in embodied intelligence, demonstrating encouraging manipulation capabilities across a range of robotic tasks, such as stacking blocks, opening drawers, and organizing household objects. The prevailing approach is to learn a reactive end-to-end policy that directly maps high-level goals and perceptual inputs to low-level motor control commands \cite{li2025comprehensive, zhang2025pure, wang2025tmcir, sapkota2025vision, guan2025efficient, wang2025vla, li2025vla,  wang2025vision}. However, this paradigm faces a critical bottleneck: learning such a holistic “perception-to-action” mapping is inherently challenging and typically requires large amounts of high-quality demonstration data \cite{ma2024survey, liu2025aligning}.

To tackle the challenges of learning a direct perception-to-action mapping.
A widely explored direction is to equip VLA models with the ability to ``think before acting'', typically instantiated through Chain-of-Thought (CoT) reasoning \cite{yin2025deepthinkvla, ye2025vla, zhang2025reasoning, guo2025vla}. Specifically, before producing actions, the model explicitly analyzes the task goal, current visual observations, object relationships, and potential subgoals, and generates a Chain-of-Thought (CoT) reasoning trace. Such reasoning enables the model to decompose high-level instructions into a sequence of executable intermediate decisions and to dynamically adjust its action strategy based on environmental feedback.

\begin{figure*}[t!]
    \centering
    \includegraphics[width=0.99\linewidth]{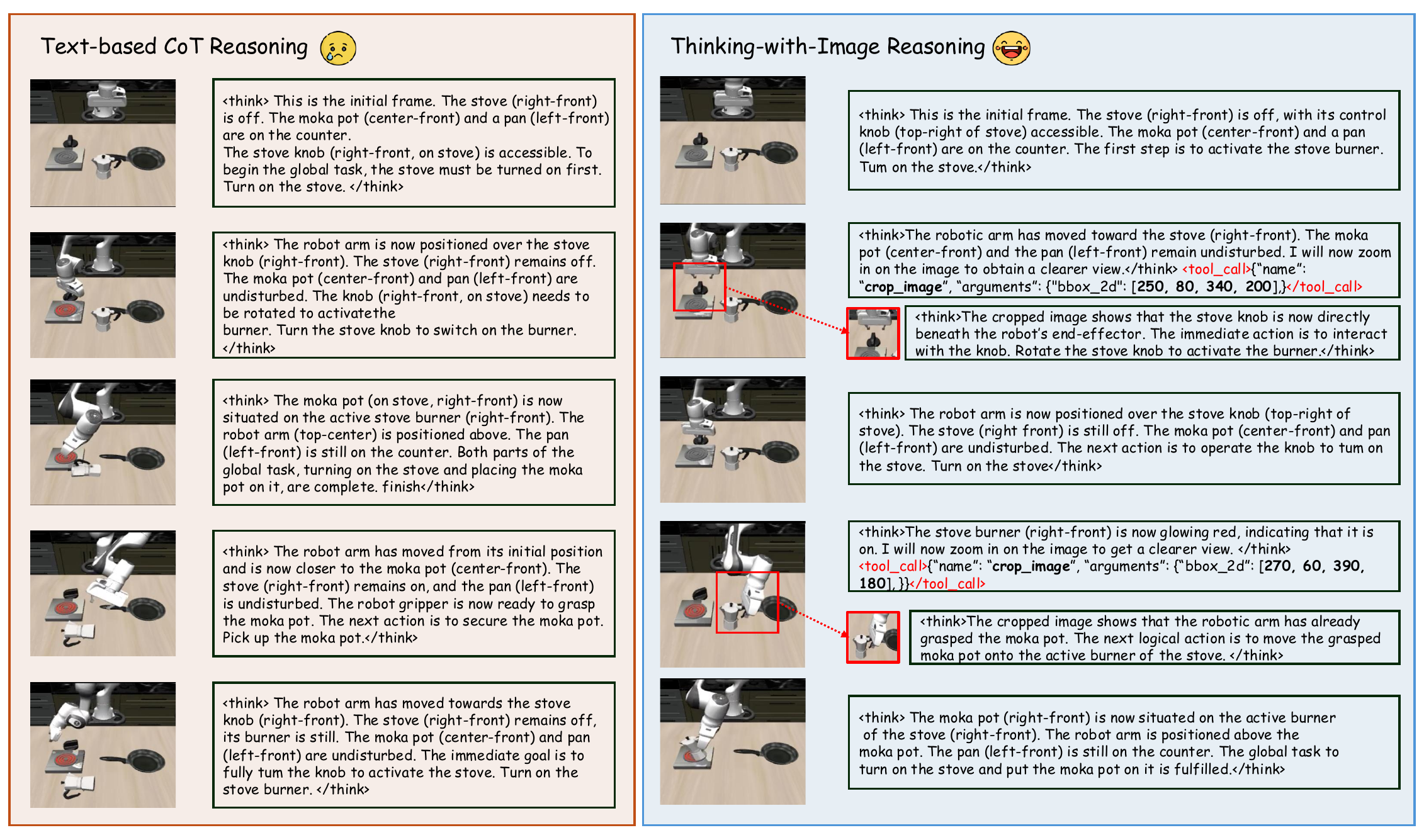}
    \caption{Comparison between text-based CoT Reasoning (left) and Thinking-with-Image Reasoning (right) for VLA.
    Left: Conventional VLA reasoning models adopt a text-based Chain-of-Thought reasoning paradigm, treating visual inputs as static context, which fails to successfully grasp the target object.
Right: Our proposed thinking-with-image framework models perception as a dynamically invocable reasoning action, enabling the model to call visual tools during intermediate reasoning steps and realize an interleaved perception–reasoning–action process, ultimately completing the manipulation task successfully.
}
    \label{fig:intro}
    \vspace{-0.45cm}
\end{figure*}

However, existing VLA reasoning models remain constrained by a text-based reasoning paradigm. In such approaches, visual inputs are encoded once into static embeddings and treated as fixed context throughout the reasoning process. Consequently, reasoning unfolds primarily in the language space, while perception becomes a passive, one-shot observation, as illustrated in Fig. \ref{fig:intro} (left). This design departs significantly from human cognitive processes \cite{sandini1992vision, bajcsy2018revisiting}, where visual perception is active, iterative, and tightly coupled with reasoning. Humans dynamically revisit the environment, selectively attend to task-relevant regions, and adapt visual focus when uncertainty arises. In contrast, static visual encoding limits a model’s ability to resolve ambiguities, track subgoals, and recover from intermediate execution errors, particularly in long-horizon manipulation tasks.

To address these challenges, we propose VLA-Thinker, a thinking-with-image reasoning framework for embodied intelligence. 
To the best of our knowledge, it is the first VLA model capable of thinking--with--image reasoning.
VLA-Thinker models perception as an explicit, dynamically invocable reasoning action. During reasoning process, the model can actively request task-relevant visual information (Relevant sub-image) through \textbf{tool invocation}, enabling perception to be interleaved with reasoning steps and action generation. This design transforms the traditional perception--reasoning--action pipeline into a tightly coupled and cooperative process, allowing the model to adapt its visual observations based on evolving reasoning needs, as illustrated in Fig. \ref{fig:intro} (right). 
  Realizing such a perception-driven reasoning approach requires the model to learn not only structured reasoning patterns, but also when and how to query the environment effectively. To this end, we introduce a two-stage training strategy.
  First, a cold-start phase leverages carefully curated visual Chain-of-Thought data to distill foundational reasoning patterns and establish consistent operation formats for perception-driven reasoning. Second, we employ Group Relative Policy Optimization (GRPO) to perform causal alignment over complete reasoning–action trajectories, encouraging the model to generate effective perception queries and actions that jointly lead to task success.

We evaluate VLA-Thinker on two representative embodied intelligence benchmarks: LIBERO \cite{LIBERO-2023} and RoboTwin 2.0 \cite{chen2025robotwin}. Experimental results demonstrate that VLA-Thinker achieves significant performance improvements on both benchmarks. In particular, VLA-Thinker attains a 97.5\% success rate on the LIBERO benchmark, representing a 6.5\% improvement over the backbone model OpenVLA-OFT \cite{OpenVLA-OFT-2025}, thereby validating the effectiveness of the proposed method.
In summary, our contributions are threefold:
\begin{itemize}
    \item  We introduce \textbf{VLA-Thinker}, the first VLA model capable of \textbf{thinking--with--image reasoning}, which models visual perception as a dynamically invocable reasoning action, enabling Multimodal Embodied Chain-of-Thought. 
    \item  We propose a two-stage training framework combining SFT cold-start and GRPO-based trajectory-level alignment, which stabilizes multimodal reasoning behaviors and effectively optimizes long-horizon reasoning–action trajectories under sparse rewards. 
    \item  Extensive experiments on multiple embodied benchmarks (LIBERO and RoboTwin 2.0) show the effectiveness of our proposed approach.
    Notably, \textbf{VLA-Thinker} achieves an average success rate of \textbf{97.5\%} on the LIBERO benchmark.
\end{itemize}

\section{Method}
\label{sec:method}

In this section, we present VLA-Thinker, the first thinking-with-image reasoning framework that tightly couples perception, reasoning, and action in embodied environments. Our method is built upon two key components. First,  we reformulate VLA reasoning as an iterative multimodal interleaved process, where visual perception is treated as a dynamically invocable reasoning action rather than a static context. This design enables the model to actively query task-relevant visual evidence during intermediate reasoning steps and generate coherent reasoning–action trajectories (Sec. \ref{sec:problem_formulation}). Second,  we introduce a two-stage training strategy consisting of (1) a supervised fine-tuning (SFT) cold-start stage that activates structured reasoning and tool-use behaviors using synthesized embodied Chain-of-Thought data, and (2) a trajectory-level reinforcement learning stage based on Group Relative Policy Optimization (GRPO), which aligns complete reasoning–action trajectories with sparse task-level success signals (Sec. \ref{sec:training}). Together, these components enable VLA-Thinker to perform robust long-horizon reasoning and grounded action execution.

\subsection{Problem Formulation}
\label{sec:problem_formulation}

\begin{figure*}[t!]
    \centering
    \includegraphics[width=0.99\linewidth]{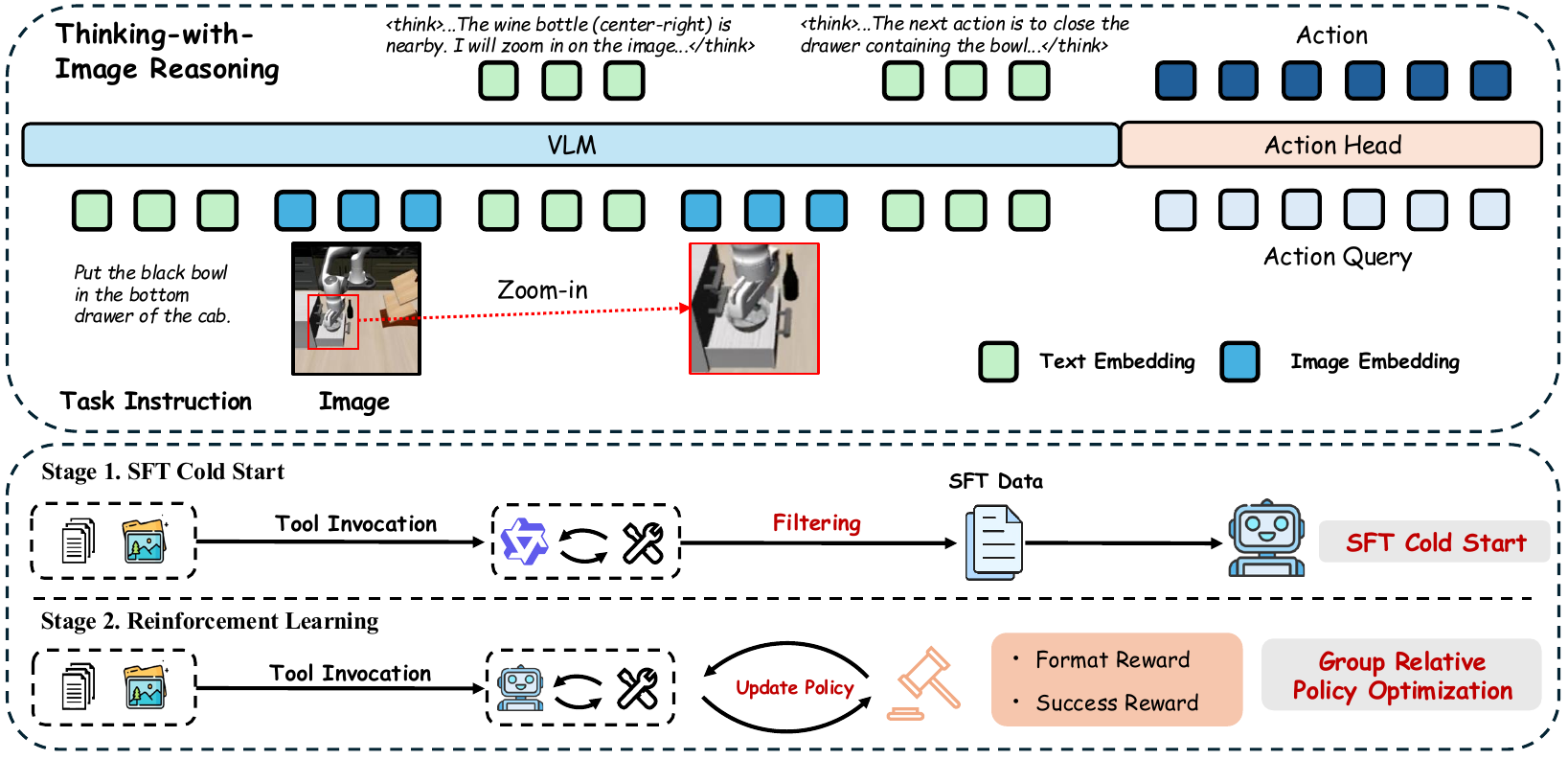}
    \caption{The upper panel illustrates the main process of our proposed \textbf{Thinking-with-Image framework}. Language instructions and visual observations are encoded into a shared VLM, enabling interleaved reasoning and dynamic zoom-in perception before action generation. The lower panel presents the two-stage training strategy: (1) SFT cold-start to activate structured reasoning and tool-use behaviors, and (2) GRPO-based reinforcement learning to align multimodal reasoning–action trajectories with task-level objectives under sparse rewards.
}
    \label{fig:method}
    \vspace{-0.45cm}
\end{figure*}

We study Vision-Language-Action (VLA) reasoning \cite{ zhao2025cot, guo2025vla} in embodied environments, where a model must generate action decisions by jointly reasoning over language instructions and visual observations. Beyond conventional formulations that treat visual inputs as static context, we introduce a \emph{thinking-with-image} reasoning paradigm that explicitly interleaves reasoning with  visual perception.

We formalize VLA thinking-with-image reasoning as an iterative \emph{multimodal interleaved reasoning process}, in which perception is modeled as an explicit reasoning operation rather than a passive input.

Given an initial language instruction $T_0$ and an initial visual observation set $V_0$ (e.g., egocentric RGB images), a VLA model iteratively produces a sequence of outputs:
\begin{equation}
A_k = f_{\text{VLA}}\big(\{T_i, C_i, V_i\}_{i=0}^{k}\big)
\end{equation}
where,
$T_k$ denotes a textual reasoning step, representing the model's intermediate hypothesis or thought,
$C_k$ denotes a perception invocation, specifying a visual tool invocation,
$V_k$ denotes the visual evidence returned by executing the perception tool.
$A_k$denotes the action generated by the model.

At each iteration, a controller (or parser) determines whether the model should:
generate the next reasoning step and perception request $(T_{k+1}, C_{k+1})$, or
terminate the reasoning process and output an environment action $A$.
If a perception action is invoked, the corresponding visual tool is executed and returns new visual evidence $V_{k+1}$, which is appended to the reasoning context and used to guide subsequent reasoning and action generation.
This process yields a \emph{multimodal reasoning--action trajectory}:
\begin{equation}
\tau = \{T_1, C_1, V_1, T_2, C_2, V_2, \dots, T_k, A_k\},
\end{equation}
where $A_k$ denotes the final environment action executed by the model.

In this work, we consider one type of visual tool: ZOOM-IN, which is used to inspect fine-grained details within a specified region of the target image. The primary objective of this study is to validate the fundamental effectiveness of the interleaved perception-reasoning-action paradigm. Therefore, we employ the zoom-in mechanism as a representative instance to verify the end-to-end pipeline and demonstrate its potential for boosting VLA performance. We anticipate that this work will serve as a baseline, and we look forward to the community exploring more diverse and sophisticated visual tools in future follow-up research.
Detailed protocols for the visual tool are provided in the Appendix.

\subsection{Training Strategies}
\label{sec:training}
We train VLA-Thinker using a two-stage training pipeline that first equips the model with foundational reasoning capabilities and subsequently aligns these capabilities with optimal task-level objectives, as illustrated in Fig. \ref{fig:method}.
\\

\noindent\textbf{Reasoning Activation via SFT Cold-Start.}
We begin with a cold-start supervised fine-tuning (Supervised Fine-Tuning, SFT) stage \cite{chen2026v, wang2025adatooler, wang2025knowing} to activate the model’s foundational reasoning capabilities and tool-use behaviors. However, existing large-scale embodied intelligence datasets generally lack explicitly annotated Chain-of-Thought (CoT) reasoning trajectories, which substantially limits effective supervision of reasoning processes.
To address this critical data gap, we leverage Qwen3-VL-30B-A3B-Instruct \cite{Qwen3-VL} to synthesize high-quality embodied CoT data. The generated reasoning trajectories include not only structured intermediate reasoning steps but also explicit modeling of valid and task-relevant tool invocation patterns.
Specifically, we first identify semantically meaningful keyframes within each trajectory by detecting changes in the gripper state. Such state transitions typically correspond to subtask boundaries, enabling an effective decomposition of embodied tasks into hierarchical structures. For these keyframes, we employ Qwen3-VL-30B-A3B-Instruct  to generate complete CoT annotations, including justified tool invocation and corresponding textual reasoning.
For the remaining intermediate frames that are not selected as keyframes, we further use Qwen3-VL-30B-A3B-Instruct  to generate pure textual CoT reasoning annotations, ensuring reasoning continuity throughout the entire trajectory.
To guarantee the reliability and consistency of the synthesized data, we enforce strict structured format validation (schema checks) and impose temporal consistency constraints on all generated annotations. Through this process, we construct a unified, clean, and high-quality embodied CoT dataset, which provides a solid foundation for stable and effective SFT training.
\\

\noindent\textbf{Learning Reasoning and Action via RL.}
After activating structured reasoning and valid tool-use behaviors via the SFT cold-start stage, we further optimize VLA-Thinker using reinforcement learning \cite{li2025simplevla, shao2024deepseekmath} to align complete reasoning--action trajectories with task-level objectives. Different from conventional action-only policy learning \cite{zang2025rlinf, li2025vla}, our goal is to jointly optimize \emph{reasoning steps, perception invocations, and environment actions} under sparse and delayed rewards.
We model VLA-Thinker as a stochastic policy $\pi_\theta$ over multimodal reasoning--action trajectories. Given an instruction $T_0$ and initial visual observation $V_0$, the policy generates a trajectory
\begin{equation}
\tau = \{T_1, C_1, V_1, \dots, T_n, A_n\},
\end{equation}
where $T_k$ denotes a textual reasoning step, $C_k$ a perception tool invocation, $V_k$ the returned visual sub-image content, and $A_n$ the final executable action.

The reward function $R(\tau)$ is sparse and is assigned only at the end of a trajectory based on a verifiable task completion signal $I_{\text{success}}$. 
No intermediate rewards are provided for the semantic correctness of the reasoning process itself. 
In addition, we introduce a small format regularization reward $I_{\text{format}}$ to prevent drift in the reasoning style. The reward function is defined as:
\begin{equation}
R(\tau) = \alpha_s \cdot I_{\text{success}} + \alpha_f \cdot I_{\text{format}} .
\end{equation}
where $\alpha_s$ and $\alpha_f$ are weighting coefficients.
Here, $I_{\text{success}} = 1$ if the task is successfully completed and $0$ otherwise, while $I_{\text{format}} = 1$ if the CoT reasoning follows the correct format (<think></think><tool></tool>) and $0$ otherwise.

For each instruction and initial observation, we sample a group of $M$ trajectories:
\begin{equation}
\{\tau_1, \tau_2, \dots, \tau_M\} \sim \pi_\theta.
\end{equation}
Given their rewards $\{R(\tau_i)\}_{i=1}^M$, we compute the relative advantage for each trajectory as:

\begin{equation}
A_i = \frac{R(\tau_i) - \text{mean}(\{R(\tau_1), R(\tau_2), \dots, R(\tau_M)\})}{\text{std}(\{R(\tau_1), R(\tau_2), \dots, R(\tau_M)\})}
\end{equation}
Following DeepSeek R1 \cite{guo2025deepseek},
the training objective is defined as:

\begin{equation}
\begin{aligned}
    &\mathcal{J}(\theta) = 
    \mathbb{E}_{q, \{\tau_i\}} \Bigg[ \frac{1}{M} \sum_{i=1}^M \Big( 
     \min \Big( \frac{\pi_\theta(\tau_i | q)}{\pi_{\theta_{\text{old}}}(\tau_i | q)} A_i, \\ &\text{clip}\Big( \frac{\pi_\theta(\tau_i | q)}{\pi_{\theta_{\text{old}}}(\tau_i | q)}, 1 - \epsilon, 
    1 + \epsilon \Big) A_i \Big) - \beta \, \mathbb{D}_{\mathrm{KL}}(\pi_\theta \| \pi_{\mathrm{ref}}) \Big) \Bigg]
\end{aligned}
\end{equation}

This relative formulation eliminates the need for an explicit value function and substantially reduces variance when optimizing long-horizon reasoning trajectories with sparse feedback.
By optimizing $\mathcal{J}(\theta)$, the VLA model is able to simultaneously enhance its reasoning capability and action execution capability, unifying both under the core objective of maximizing final task success.

\subsection{Discussion}

The core philosophy of VLA-Thinker is to transition from a “passive observation” paradigm to an “active perception–reasoning” paradigm. Traditional VLA models typically treat vision as a static, one-shot input, thereby decoupling perception from the subsequent multi-step reasoning process. In contrast, by modeling perception as a dynamically invocable reasoning action, our framework enables the model to actively revisit the environment to resolve ambiguities and recover from execution errors. The synergy between our two-stage training strategy is crucial to realizing this philosophy. First, the SFT cold-start phase does more than teach the model to “verbalize” reasoning; it establishes fundamental causal links between specific visual uncertainties and the necessity of tool invocation, injecting structured priors into the policy. Subsequently, GRPO reinforces complete (thought, tool, action) trajectories using task-level success signals, effectively optimizing not only “how to reason” but also “when to invoke.” Through this process, the model learns to balance reasoning cost against task success, ultimately learning when not to think to avoid redundant computation. Although this work primarily employs ZOOM-IN as the visual tool, our fundamental objective is to validate the effectiveness and feasibility of the interleaved reasoning paradigm within the VLA architecture. The zoom-in mechanism serves as a representative instance to instantiate the full pipeline and demonstrate its potential to enhance decision robustness in complex manipulation. We believe that the primary contribution of VLA-Thinker lies in this extensible framework rather than in any specific tool design.

\section{Experiment}
\label{sec:exp}
\subsection{Experimental Setup}


\begin{table}[!h]
  \centering
  \caption{RoboTwin 2.0 task classification based on planning horizon and required steps.}
  \resizebox{.99\linewidth}{!}{
  \begin{tabular}{l|c|c|c}
    \toprule
    \textbf{Task Name} & \textbf{Steps} & \textbf{Horizon} & \textbf{Horizon Group} \\
    \midrule
    \multicolumn{4}{c}{\cellcolor{gray!20}\textbf{Short Horizon Tasks (112-130 steps)}} \\
    \midrule
    lift\_pot & 112 & Short & \multirow{4}{*}{\makecell{Average: 121 steps\\Count: 4 tasks}} \\
    beat\_block\_hammer & 113 & Short & \\
    pick\_dual\_bottles & 127 & Short & \\
    place\_phone\_stand & 130 & Short & \\
    \midrule
    \multicolumn{4}{c}{\cellcolor{gray!20}\textbf{Medium Horizon Tasks (151-223 steps)}} \\
    \midrule
    move\_can\_pot & 151 & Medium & \multirow{4}{*}{\makecell{Average: 176 steps\\Count: 4 tasks}} \\
    place\_a2b\_left & 155 & Medium & \\
    place\_empty\_cup & 174 & Medium & \\
    handover\_mic & 223 & Medium & \\
    \midrule
    \multicolumn{4}{c}{\cellcolor{gray!20}\textbf{Long Horizon Tasks (283-313 steps)}} \\
    \midrule
    handover\_block & 283 & Long & \multirow{2}{*}{\makecell{Average: 298 steps\\Count: 2 tasks}} \\
    stack\_bowls\_two & 313 & Long & \\
    \midrule
    \multicolumn{4}{c}{\cellcolor{gray!20}\textbf{Extra Long Horizon Tasks (466-637 steps)}} \\
    \midrule
    blocks\_rank\_rgb & 466 & Extra-Long & \multirow{2}{*}{\makecell{Average: 552 steps\\Count: 2 tasks}} \\
    put\_bottles\_dustbin & 637 & Extra-Long & \\
    \midrule
    \multicolumn{1}{l|}{\textbf{Overall Statistics}} & \multicolumn{3}{c}{\textbf{Total: 12 tasks, Average: 256 steps}} \\
    \bottomrule
  \end{tabular}
  }
  \label{tab:task_horizon_analysis}
\end{table}

\noindent\textbf{Benchmarks.}
We evaluate VLA-Thinker on the LIBERO benchmark \cite{liu2023libero} and the RoboTwin 2.0 benchmark \cite{chen2025robotwin}.
LIBERO is a language-guided manipulation benchmark designed for lifelong learning, covering diverse object types, task specifications, and environment settings. It consists of five task suites: LIBERO-Goal, LIBERO-Spatial, LIBERO-Object, LIBERO-Long (10 tasks, each with 50 expert demonstrations), and LIBERO-90 (containing 90 tasks for large-scale multi-task evaluation).
We use the average Success Rate (SR) over 50 held-out test scenes per task as the evaluation metric.
RoboTwin2.0 is a simulation benchmark for bimanual manipulation, comprising 50 dual arm collaborative tasks and covering diverse robot morphologies and 731 object instances. The benchmark incorporates comprehensive domain randomization mechanisms (clutter, lighting, background, tabletop height, and language instructions), which enhance task diversity and improve sim to real generalization and transfer capability.
During training and evaluation on RoboTwin2.0, we adopt domain randomized task settings and evaluate each task on 100 held out test scenarios. Specifically, we select 12 representative tasks and categorize them into four different temporal horizon levels based on their average execution steps, enabling a stratified and comprehensive evaluation. Tab. \ref{tab:task_horizon_analysis} summarizes the average number of steps for each task, as well as the step ranges corresponding to different horizon levels.\\

\noindent\textbf{Backbones.}
We adopt OpenVLA-OFT \cite{kim2025fine} as the base model.
The model is built upon OpenVLA \cite{kim2024openvla}, adopting a vision encoder and LLaMA2-7B \cite{touvron2023llama} as the backbone, and incorporating action chunking together with a parallel decoding design. This architecture provides high efficiency in online reinforcement learning scenarios that require frequent inference.
To improve training and inference efficiency, we use only single view images, language instructions, and robot proprioceptive states as model inputs, while the official model additionally utilizes wrist camera images. Moreover, for the LIBERO tasks, we do not use robot proprioceptive states as inputs.
In terms of model architecture, we adopt only the parallel decoding and action chunk designs.\\

\begin{table}[!t]
  \centering
  \caption{Main results of different VLA models on LIBERO. All reported values denote the task
success rate (SR, \%) evaluated under 50 randomized initial conditions per task, averaged within each suite and across all suites. 
\textbf{Bold numbers} indicate the best performance within each suite.}
  \resizebox{.99\linewidth}{!}{
    \begin{tabular}{lccccc}
      \toprule
      \multirow{2}{*}{\textbf{Model}} &
        \multicolumn{5}{c}{\textbf{LIBERO}} \\
      \cmidrule(lr){2-6}
      & \textbf{Spatial} & \textbf{Object} & \textbf{Goal} & \textbf{Long} & \textbf{Avg} \\
      \midrule
      FlowVLA \cite{FlowVLA-2025}  &93.2 &95.0 &91.6 &72.6& 88.1\\
        UnifiedVLA \cite{UnifiedVLA-2025}  &95.4 &\underline{\textit{98.8}} &93.6 &94.0 &95.5 \\
        OpenVLA \cite{OpenVLA-2024}  &84.7 &88.4 &79.2 &53.7 &76.5 \\
		UniVLA \cite{UniVLA-2025}  &96.5  &96.8  &95.6  &92.0 &95.2 \\
		CoT-VLA \cite{CoT-VLA-2025}  &87.5 &91.6 &87.6 &69.0 &81.1 \\
		WorldVLA \cite{WorldVLA-2025}  &87.6  &96.2  &83.4  &60.0 &81.8 \\	
		TraceVLA \cite{TraceVLA-2025}  &84.6 &85.2 &75.1 &54.1 &74.8 \\	
        MolmoAct \cite{MolmoAct-2025}  &87.0 &95.4 &87.6 &77.2 &86.6 \\	
        ThinkAct \cite{ThinkAct-2025}  &88.3 &91.4 &87.1 &70.9 &84.4 \\
        PD-VLA \cite{PDVLA-2025}  &95.5&96.7&94.9&91.7&94.7\\
		
		4D-VLA \cite{4D-VLA-2025} &88.9&95.2& 90.9& 79.1& 88.6\\
		SpatialVLA \cite{SpatialVLA-2025} &88.2 &89.9 &78.6 &55.5 &78.1 \\
		$\pi_0$ \cite{pi0-2025} &96.8 &\underline{\textit{98.8}} &95.8 &85.2 &94.2 \\
		$\pi_0$-FAST \cite{pi0-FAST-2025} &96.4 &96.8 &88.6 &60.2 &85.5 \\
        NORA \cite{NORA-2025} &92.2 &95.4 &89.4 &74.6 &87.9 \\
		SmolVLA \cite{SmolVLA-2025} &93.0 &94.0 &91.0 &77.0 &88.8 \\	
		GR00T N1 \cite{GR00T-N1-2025} &94.4 &97.6 &93.0 &90.6 &93.9\\
        GraspVLA \cite{GraspVLA-2025} & - & 94.1&91.2&82.0 &89.1\\
		
		Seer$^{\dag}$ \cite{Seer-2025}   & - & - & - &78.7 & 78.7 \\
		VLA-OS \cite{VLA-OS-2025} &87.0& 96.5 &92.7 &66.0 &85.6\\
		Diffusion Policy$^{\dag}$ \cite{DP-2023} &78.3 &92.5 &68.3 &50.5 &72.4 \\	
      \midrule
      OpenVLA-OFT         & 91.6 & 95.3 & 90.6 & 86.5 & 91.0 \\
      \textbf{VLA-Thinker (Ours)}       & \textbf{98.7} & \textbf{99.0} & \textbf{95.2} & \textbf{96.9} & \textbf{97.5} \\
      \rowcolor[rgb]{ .900,  .900,  .900} \quad $\Delta$ &
        \textcolor{red}{$+7.1$} &
        \textcolor{red}{$+3.7$} &
        \textcolor{red}{$+4.6$} &
        \textcolor{red}{$+10.4$} &
        \textcolor{red}{$+6.5$} \\
      \bottomrule
    \end{tabular}
  }
  \label{exp:libero}
\end{table}

\noindent\textbf{Implementation Details.}
VLA-Thinker is initialized from the publicly available OpenVLA-OFT weights \cite{kim2025fine} and is trained on 8 NVIDIA H100 GPUs.
During training and inference, we only use single-view images, language instructions, and robot proprioceptive states as model inputs, whereas the official OpenVLA-OFT model \cite{OpenVLA-OFT-2025} additionally incorporates wrist camera images.
Besides, in the LIBERO, we don’t include robot
proprioceptive states in model inputs.
The model is trained using a two-stage pipeline. In the first stage, a cold-start supervised fine-tuning (SFT) is performed on our constructed embodied Chain-of-Thought (CoT) dataset.
In the second stage, we further introduce online reinforcement learning (online RL) to align the generated CoT reasoning with downstream action execution. This stage leverages outcome-based reward signals and is trained using the GRPO reinforcement learning algorithm. 
The batch size is set to 64 during the SFT stage and 128 during the RL stage. The learning rate is configured as $1 \times 10^{-5}$ for SFT and $2 \times 10^{-6}$ for RL, and both stages are optimized using the AdamW optimizer \cite{loshchilov2017decoupled}.
Overall, the complete training process takes approximately 3 days.
Additional training details, including dataset statistics, hyperparameter configurations, and inference settings, are provided in Appendix.

\begin{table}[!t]
  \centering
  \caption{Main results of different VLA models on RoboTwin2.0. All reported values denote the task
success rate (SR, \%) evaluated under 100 randomized initial conditions per task. 
\textbf{Bold numbers} indicate the best performance within each task.}
  \label{exp:Robotwin2.0}
  \resizebox{\linewidth}{!}{
    \begin{tabular}{lccccc}
      \toprule
      \multicolumn{6}{c}{\textbf{Short Horizon Tasks (100-130 Steps)}} \\
      \midrule
      \textbf{Model} & \textbf{\scalebox{0.9}{Lift Pot}} & \textbf{\scalebox{0.9}{Beat Hammer Block}} & \textbf{\scalebox{0.9}{ Pick Dual Bottles}} & \textbf{\scalebox{0.9}{Place Phone Stand}} & \textbf{Avg} \\
      \midrule
      $\pi_0$ \cite{pi0-2025} & 51.0 & 59.0 & 50.0 & 22.0 & 45.5 \\
      RDT \cite{RDT-2025} & 45.0 & 22.0 & 18.0 & 13.0 &  24.5 \\
      $\pi_0$-FAST \cite{pi0-FAST-2025} & 30.0 & 38.0 & 25.0 & 16.0 & 27.3 \\
      DeepThinkVLA \cite{yin2025deepthinkvla} & 62.0 & 73.0 & 61.0 & 24.0 & 55.0 \\
      \midrule
      OpenVLA-OFT \cite{OpenVLA-OFT-2025} & 10.1 & 28.1 & 29.7 & 17.1 & 21.3 \\
      \textbf{VLA-Thinker (Ours)} & \textbf{64.8} & \textbf{82.5} & \textbf{65.4} & \textbf{36.6} & \textbf{62.3} \\
      \rowcolor[rgb]{ .900,  .900,  .900} \quad $\Delta$ & \textcolor{red}{$+54.7$} & \textcolor{red}{$+54.4$} & \textcolor{red}{$+35.7$} & \textcolor{red}{$+19.5$} & \textcolor{red}{$+41.0$} \\
      \midrule
      \multicolumn{6}{c}{\textbf{Medium Horizon Tasks (150-230 Steps)}} \\
      \midrule
      \textbf{Model} & \textbf{\scalebox{0.9}{Move Can Pot}} & \textbf{\scalebox{0.9}{Place A2B Left}} & \textbf{\scalebox{0.9}{Place Empty Cup}} & \textbf{\scalebox{0.9}{Handover Mic}} & \textbf{Avg} \\
      \midrule
      $\pi_0$ \cite{pi0-2025} & 41.0 & 38.0 & 60.0 & 96.0 & 58.8 \\
      RDT \cite{RDT-2025} & 33.0 & 21.0 & 42.0 & 95.0 &  47.8 \\
      $\pi_0$-FAST \cite{pi0-FAST-2025} & 34.0 & 36.0 & 54.0 & 83.0 & 51.8 \\
      DeepThinkVLA \cite{yin2025deepthinkvla} & 52.0 & 38.0 & 83.0 & 88.0 & 65.3 \\
      \midrule
      OpenVLA-OFT \cite{OpenVLA-OFT-2025} & 28.1 & 37.5 & 77.3 & 45.3 & 47.1 \\
      \textbf{VLA-Thinker (Ours)} & \textbf{61.0} & \textbf{39.1} & \textbf{92.7} & \textbf{89.9} & \textbf{70.7} \\
      \rowcolor[rgb]{ .900,  .900,  .900} \quad $\Delta$ & \textcolor{red}{$+32.9$} & \textcolor{red}{$+1.6$} & \textcolor{red}{$+15.3$} & \textcolor{red}{$+44.6$} & \textcolor{red}{$+23.6$} \\
      \midrule
      \multicolumn{6}{c}{\textbf{Long (280-320 Steps) \& Extra Long Horizon Tasks (450-650 Steps)}} \\
      \midrule
      \textbf{Model} & \textbf{\scalebox{0.9}{Handover Block}} & \textbf{\scalebox{0.9}{Stack Bowls Two}} & \textbf{\scalebox{0.9}{Blocks Rank Rgb}} & \textbf{\scalebox{0.9}{Put Bottles Dustbin}} & \textbf{Avg} \\
      \midrule
      $\pi_0$ \cite{pi0-2025} & 39.0 & 53.0 & 45.0 & 36.0 & 43.3 \\
      RDT \cite{RDT-2025} & 26.0 & 42.0 & 17.0 & 26.0 &  27.8 \\
      $\pi_0$-FAST \cite{pi0-FAST-2025} & 32.0 & 48.0 & 28.0 & 27.0 & 33.8 \\
      DeepThinkVLA \cite{yin2025deepthinkvla} & 43.0 & 62.0 & 77.0 & 49.0 & 57.8 \\
      \midrule
      OpenVLA-OFT \cite{OpenVLA-OFT-2025} & 33.1 & 40.6 & 70.2 & 42.2 & 46.5 \\
      \textbf{VLA-Thinker (Ours)} & \textbf{52.8} & \textbf{71.1} & \textbf{79.3} & \textbf{55.4} & \textbf{64.6} \\
      \rowcolor[rgb]{ .900,  .900,  .900} \quad $\Delta$ & \textcolor{red}{$+19.7$} & \textcolor{red}{$+30.5$} & \textcolor{red}{$+9.1$} & \textcolor{red}{$+13.2$} & \textcolor{red}{$+18.1$} \\
      
      \bottomrule
    \end{tabular}
  }
  \label{exp:robotwin2.0}
\end{table}

\subsection{Main Results}
\noindent\textbf{LIBERO Benchmark Comparison.}
We evaluate VLA-Thinker on the LIBERO benchmark, which consists of four language-conditioned manipulation suites (Spatial, Object, Goal, and Long), covering diverse structured reasoning and long-horizon control scenarios. As shown in Tab. \ref{exp:libero}, VLA-Thinker achieves 98.7\%, 99.0\%, 95.2\%, and 96.9\% success rates on the four suites, respectively, yielding an overall average of 97.5\%, which establishes a new state-of-the-art performance among all compared VLA models. Compared with the strong OpenVLA-OFT baseline (91.0\% Avg.), our method achieves a substantial +6.5\% overall improvement, with particularly pronounced gains on the Spatial (+7.1\%) and Long (+10.4\%) suites. These improvements indicate that explicitly modeling perception as a dynamically invocable reasoning action significantly enhances spatial grounding and long-horizon stability.  The strong performance across all four suites suggests that integrating perception into the reasoning loop leads to more robust subgoal tracking, better ambiguity resolution, and improved action consistency under complex task specifications.\\

\noindent\textbf{RoboTwin2.0 Benchmark Comparison.}
We further evaluate VLA-Thinker on RoboTwin 2.0, a challenging dual-arm manipulation benchmark characterized by strong domain randomization and extended planning horizons. As summarized in Tab. \ref{exp:robotwin2.0}, VLA-Thinker achieves 62.3\% average success on short-horizon tasks (100–130 steps), outperforming $\pi_0$ (45.5\%), DeepThinkVLA (55.0\%), and OpenVLA-OFT (21.3\%) by large margins. On medium-horizon tasks (150–230 steps), performance increases to 70.7\%, exceeding DeepThinkVLA by over 5\% and OpenVLA-OFT by more than 20\%, demonstrating improved stability under moderate planning complexity. For long and extra-long horizon tasks (280–650 steps), VLA-Thinker achieves 64.6\% average success, with notable gains on tasks such as Handover Block and Stack Bowls Two. Importantly, the relative performance advantage becomes more significant as task horizon increases, suggesting that thinking-with-image reasoning effectively mitigates error accumulation in long reasoning–action chains. By dynamically revisiting the environment and selectively querying task-relevant visual evidence, the model maintains coherent subgoal progression and exhibits stronger recovery capability when intermediate execution deviations occur. These results collectively validate that integrating active perception into the reasoning process is particularly beneficial for extended dual-arm coordination and complex temporal planning scenarios.

\subsection{Ablation Study}
We conduct ablation studies to analyze (1) the contribution of thinking-with-image reasoning and (2) the effectiveness of the two-stage training pipeline.\\

\begin{table}[t!]
    \centering
    \caption{Ablation Study on training stages.}
    \label{exp:aba}
    \renewcommand{\arraystretch}{1.2}
    \resizebox{.98\linewidth}{!}{
        \begin{tabular}{lccccc}
            \toprule
            \textbf{Method} & \textbf{Spatial} & \textbf{Object} & \textbf{Goal} & \textbf{Long } & \textbf{Avg. } \\
            \midrule
            OpenVLA-OFT \cite{OpenVLA-OFT-2025}  & 91.6 & 95.3 & 90.6 & 86.5 & 91.0  \\	
        
            VLA-Thinker-SFT & 95.9 & 96.7 & 93.4 & 94.0 & 95.0 \\
            VLA-Thinker-GRPO & 90.6 & 88.5 & 87.2 & 86.7 & 88.2 \\
            \midrule
            VLA-Thinker & \textbf{98.7} & \textbf{99.0} & \textbf{95.2} & \textbf{96.9} & \textbf{97.5} \\
            \bottomrule
        \end{tabular}
    }
\end{table}

\noindent\textbf{Ablation of Thinking-with-Image Reasoning.}
To isolate the effect of dynamic perception reasoning, we compare VLA-Thinker with OpenVLA-OFT. OpenVLA-OFT is an end-to-end vision–language–action policy model that directly predicts actions without explicit intermediate reasoning. As shown in Tab. \ref{exp:aba}, introducing the thinking-with-image reasoning mechanism improves the overall LIBERO performance from 91.0\% to 97.5\%. The gains are particularly pronounced in the Spatial and Long suites, where precise spatial grounding and consistent subgoal tracking are critical.
This comparison indicates that single-pass static visual encoding and direct action mapping are limited when handling fine-grained ambiguities and long-horizon tasks. In contrast, VLA-Thinker models perception as an explicitly invocable intermediate operation, enabling the model to query additional visual evidence through tool calls during decision-making. This leads to more grounded action selection under uncertainty. The performance improvements suggest that incorporating perception into the reasoning loop enhances policy robustness in complex scenarios and establishes a tighter perception–reasoning–action coupling, rather than merely improving visual representation quality.\\

\noindent\textbf{Ablation of Training Pipeline.}
As shown in Tab. \ref{exp:aba}, we further investigate the contribution of each training stage by evaluating variants trained with only SFT cold-start or only GRPO reinforcement learning. The SFT-only model achieves 95.0\% average success, demonstrating that structured CoT supervision effectively activates reasoning format, tool-use patterns, and multimodal interaction behavior. However, without trajectory-level reinforcement alignment, the model does not fully optimize reasoning for final task success. Conversely, directly applying GRPO without SFT initialization results in severe performance degradation (88.2\%), highlighting the instability of sparse-reward RL when structured reasoning priors are absent. These findings confirm that SFT provides essential inductive biases and stabilizes reasoning behavior, while GRPO performs causal alignment over complete reasoning–action trajectories. The combination of both stages achieves the best performance (97.5\%), demonstrating that reasoning activation and trajectory-level optimization are complementary and jointly indispensable for effectively training thinking-with-image VLA policies.

\subsection{Training Curves}

\begin{figure*}[t!]
    \centering
    \begin{subfigure}[b]{0.45\textwidth}
        \centering
        \includegraphics[width=\textwidth]{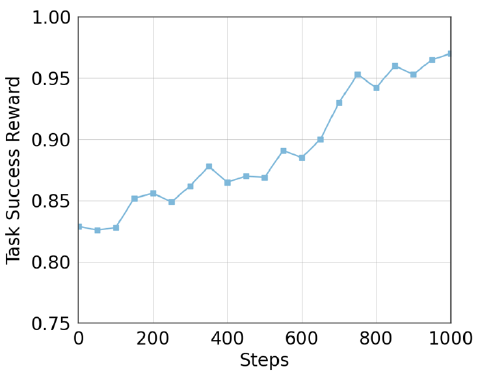}
        \vspace{-1.7em}
        \caption{Task Success Reward}
        \label{fig:succ_reward}
    \end{subfigure}
    \hfill
    \begin{subfigure}[b]{0.45\textwidth}
        \centering
        \includegraphics[width=\textwidth]{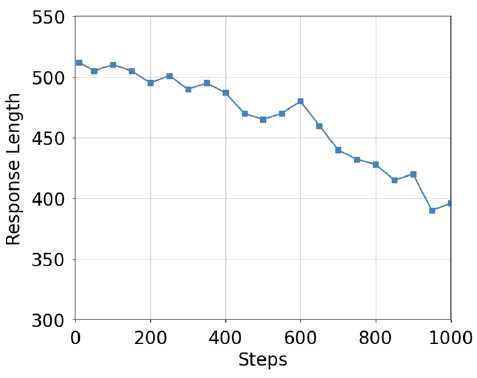}
        \vspace{-1.7em}
        \caption{Response Length}
        \label{fig:res_len}
    \end{subfigure}
    \caption{RL Training curves.
    (a) Task success reward steadily increases during GRPO training, demonstrating effective trajectory-level alignment under sparse rewards. (b) The average response length gradually decreases, indicating that the policy learns to invoke visual tools more selectively and reduce redundant reasoning.}
    \label{fig:training_curves}
    \vspace{-10pt}
\end{figure*}

Fig. \ref{fig:training_curves} illustrates the RL training dynamics of VLA-Thinker. 
As shown in Fig. \ref{fig:succ_reward}, the task success reward exhibits a clear upward trend throughout training. Starting from an initial success level of approximately 0.82, the reward steadily increases and eventually converges near 0.96. This consistent improvement demonstrates that GRPO effectively aligns the multimodal reasoning–action trajectories with the final task objective under sparse reward supervision. Importantly, the improvement is gradual rather than abrupt, indicating stable trajectory-level policy updates enabled by relative advantage normalization within sampled groups.
Fig. \ref{fig:res_len} illustrates the trend of the average response length. We observe that as training progresses, the reasoning length gradually decreases.
During the SFT cold-start stage, the model is trained mainly by imitating trajectories with complete reasoning processes. As a result, it tends to invoke tools frequently, even in relatively simple scenarios where additional visual queries are unnecessary, leading to a higher number of tool calls and comparatively redundant reasoning traces. After entering the RL stage, outcome-based policy optimization progressively reshapes the model’s behavior. The model gradually learns to autonomously determine whether tool invocation is necessary based on task requirements. When critical information is missing or visual ambiguity exists, it actively requests additional visual evidence; when the current observation is sufficient for decision-making, it directly generates actions and avoids redundant tool calls. Eventually, the frequency of tool usage becomes more reasonable and stable, and the overall reasoning length correspondingly decreases.

\section{Related Work}

\noindent\textbf{Vision--Language--Action Models.}
VLA models unify perception, language understanding, and embodied action within a single framework. Early efforts such as SayCan \cite{brohan2023can} grounded large language models in robotic affordances, while Gato~\cite{reed2022generalist} and RT-1~\cite{RT-1-2023} explored generalist multi-task transformers trained on large-scale demonstrations. PaLM-E~\cite{driess2023palm} embedded continuous sensor modalities into a large language model for embodied reasoning.
Subsequent work has broadened accessibility and scalability: the Open X--Embodiment project~\cite{OpenX-2024} assembled a cross-embodiment dataset spanning dozens of robot types, and Octo~\cite{Octo-2024} and OpenVLA~\cite{OpenVLA-2024} released open-source generalist policies efficiently fine-tunable to new robots. To improve action generation, $\pi_0$~\cite{pi0-2025} replaced autoregressive decoding with flow matching for high-frequency dexterous manipulation. 
Most recently, VLAs have been scaled to humanoid platforms: Helix~\cite{Helix-2025} introduced a dual-system architecture for full upper-body humanoid control, GR00T N1~\cite{GR00T-N1-2025} combined a vision-language backbone with a diffusion-transformer action head trained on heterogeneous data.
Unlike prior VLA models that primarily focus on architectural scaling or action representation improvements, VLA-Thinker redefines the role of perception. Instead of treating visual inputs as static context, we model perception as a dynamically invocable reasoning action, enabling interleaved perception–reasoning–action.\\

\noindent\textbf{VLA Reasoning.}
Standard VLAs learn a direct observation-to-action mapping without intermediate reasoning, limiting generalization in complex, long-horizon tasks. To address this, recent work injects structured reasoning into VLAs through supervised fine-tuning (SFT): ECoT~\cite{duan2025fast} introduces textual chain-of-thought reasoning with automatically generated annotations covering plans, sub-tasks, and visual grounding before action prediction; CoT-VLA~\cite{CoT-VLA-2025} instead generates subgoal images as a visual reasoning step leveraging action-free video data; RoboBrain~\cite{ji2025robobrain} and Robix~\cite{fang2025robix} further construct spatio-temporal thought-trace datasets with reinforced fine-tuning to strengthen causal reasoning.
In the meantime, inspired by Large Reasoning Models, a parallel line of work applies reinforcement learning (RL) to enhance embodied reasoning: Robot-R1~\cite{kim2025robot} and Embodied-R1~\cite{yuan2025embodied} use GRPO to reinforce VLM-based spatial reasoning, with the former surpassing GPT-4o at only 7B parameters and the latter achieving 87.5\% zero-shot success on real-world robotic tasks.
VLA-RL~\cite{lu2025vla} and SimpleVLA-RL~\cite{li2025simplevla} apply online RL directly to auto-regressive VLA policies via trajectory-level formulations and scalable parallelization, attaining state-of-the-art performance on LIBERO and RoboTwin. 
While existing reasoning-enhanced VLAs rely on textual CoT supervision or action-level reinforcement learning, they largely remain text-based or optimize actions independently of perception. VLA-Thinker integrates perception into the reasoning loop and performs trajectory-level GRPO alignment over complete multimodal reasoning–action sequences, enabling stable long-horizon reasoning under sparse rewards.

\section{Conclusion}
\label{sec:Con}
In this paper, we present VLA-Thinker, a thinking-with-image reasoning framework that integrates perception into the reasoning loop of VLA models. Unlike text-based CoT approaches that treat visual inputs as static context, our method models perception as a dynamically invocable reasoning action, enabling interleaved perception–reasoning–action trajectories.
We further propose a two-stage training pipeline combining SFT-based reasoning activation and GRPO-based trajectory-level alignment. Extensive experiments on LIBERO and RoboTwin 2.0 demonstrate that VLA-Thinker significantly outperforms strong baselines, achieving a 97.5\% success rate on LIBERO. These results suggest that explicitly coupling perception with reasoning is crucial for robust and long-horizon embodied decision-making.



\bibliographystyle{splncs04}
\bibliography{main}

\clearpage
\appendix


\begin{figure*}[!t]
    \centering
    \includegraphics[width=0.99\linewidth]{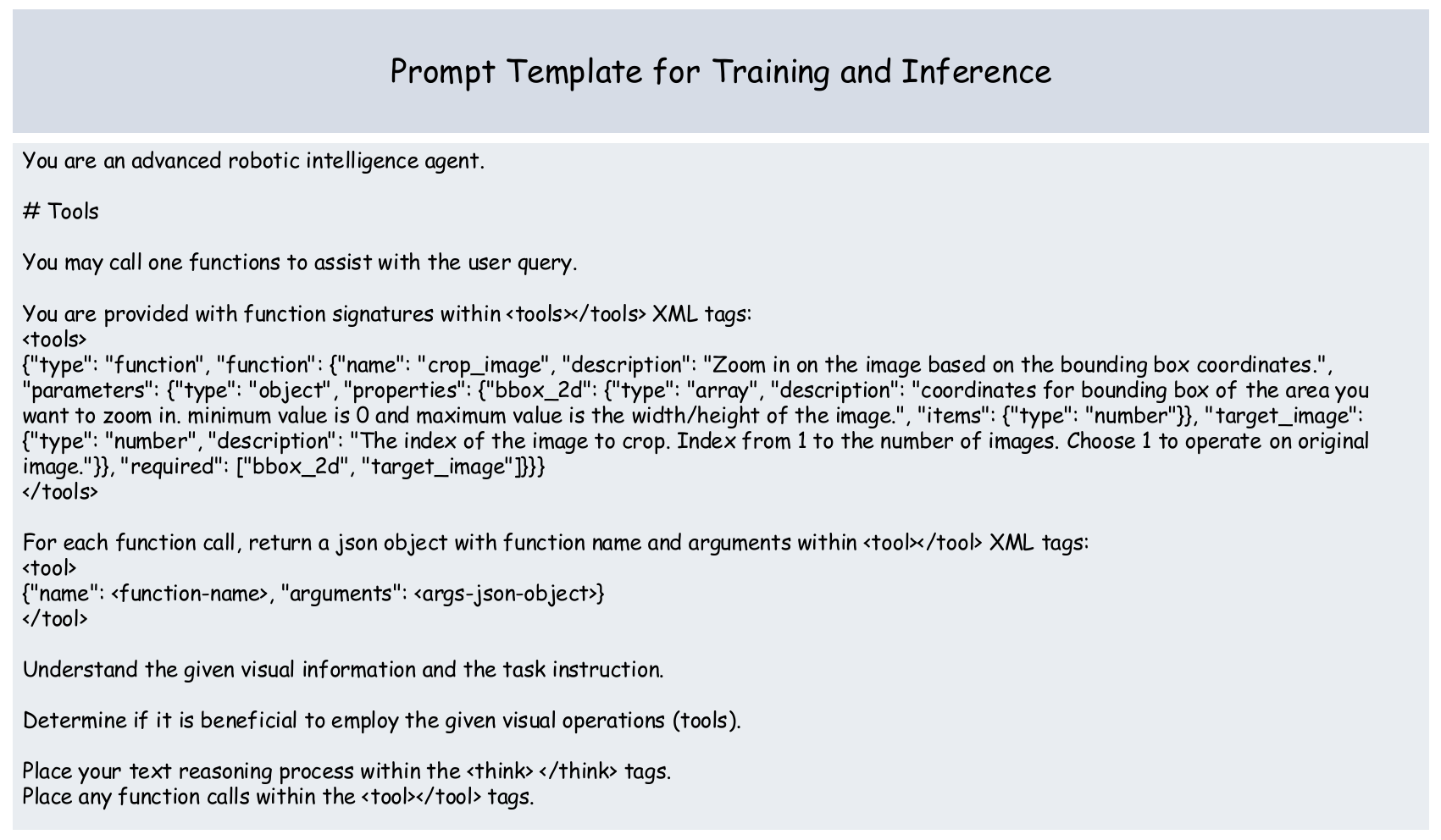}
    \caption{Prompt template for training and inference.
}
    \label{fig:prompt}
    \vspace{-0.45cm}
\end{figure*}

\section{Prompt Template}

Fig. \ref{fig:prompt} illustrates the prompt template for training and inference of VLA-Thinker.

\section{Additional Implementation Details}
\label{Imple_details}

VLA-Thinker is initialized from the public OpenVLA-OFT weight \cite{OpenVLA-OFT-2025}.

\noindent \textbf{Dataset Construction.}
Before performing Supervised Fine-Tuning (SFT), we construct two embodied Chain-of-Thought (CoT) datasets based on the public LIBERO demonstrations and Robotwin2.0 demonstrations, following the two-stage pipeline described in Section 2.2.
This process generates 273,465 annotated keyframes and 215,784 annotated keyframes, respectively, which serve as the supervision data for the cold-start stage. \\

\noindent \textbf{Supervised Fine-Tuning (SFT).}
During the SFT stage, the model is trained for 100k steps using a batch size of 64 and a learning rate of $1 \times 10^{-5}$. We employ a hybrid attention mask that enables two complementary supervision modes within a single forward pass: CoT tokens are optimized autoregressively, while action tokens are supervised bidirectionally. Model parameters are optimized using a token-level cross-entropy loss. \\

\noindent \textbf{Reinforcement Learning (RL).}
In the reinforcement learning stage, we adopt Group Relative Policy Optimization (GRPO) \cite{shao2024deepseekmath}. Each trajectory receives a sparse task-success reward, complemented by a small format-regularization reward to maintain the quality and consistency of generated CoT reasoning. Policy optimization is performed with a mini-batch size of 128, a low clip ratio $\epsilon = 0.2$, and a high clip ratio $\epsilon = 0.28$. Additionally, a KL penalty relative to the SFT reference model is introduced to mitigate catastrophic forgetting during policy updates.

\noindent \textbf{Infrastructure and Inference.}
Training is conducted on 8 NVIDIA H100 GPUs. During inference, we adopt greedy decoding for both reasoning and action tokens.

\section{Inference Speed}
We evaluate the inference efficiency of VLA-Thinker in comparison with the end-to-end OpenVLA-OFT \cite{OpenVLA-OFT-2025} on the LIBERO benchmark \cite{liu2023libero} using an H100 GPU. On average, VLA-Thinker requires 19\% more execution time than OpenVLA-OFT, primarily due to its autoregressive reasoning process.
Despite this moderate increase in inference time, the proposed embodied reasoning mechanism—serving as a form of test-time scaling—substantially improves downstream task performance. Specifically, VLA-Thinker consistently outperforms OpenVLA across all four LIBERO task categories, achieving success rate improvements of 7.1\% on spatial tasks, 3.7\% on object tasks, 4.6\% on goal tasks, and 10.4\% on long-horizon tasks.
These results indicate that the additional computational overhead introduced by reasoning is well justified by the resulting performance gains, highlighting the effectiveness of embodied reasoning for enhancing robotic manipulation capabilities.

\section{Limitations and Future Works}

Although VLA-Thinker demonstrates strong performance on both LIBERO and RoboTwin 2.0 benchmarks, several limitations remain. First, the current framework employs only a single visual tool (ZOOM-IN) to validate the effectiveness of thinking-with-image reasoning. While sufficient to demonstrate the proposed paradigm, more diverse perception tools (e.g., object grounding, segmentation, or web-search) may further enhance reasoning capability in complex environments. 
Second, since VLA-Thinker is built upon pretrained multimodal large language models (MLLMs), it inevitably inherits their inherent limitations, particularly the issue of hallucination in visual or spatial reasoning. This may cause the generated actions to reference incorrect object attributes or spatial relationships, thereby affecting the subsequent execution process. Future progress in mechanisms for mitigating MLLM hallucinations could further improve the robustness and reliability of the system for real-world deployment.


\end{document}